# SimArch: A Multi-agent System For Human Path Simulation In Architecture Design

**Yen-Chia Hsu**
Department of Architecture in Tangible Interaction Design
Carnegie Mellon University
Pittsburgh, PA 15213
*hsu.yenchia@gmail.com*

## Abstract

Human moving path is an important feature in architecture design. By studying the path, architects know where to arrange the basic elements (e.g. structures, glasses, furniture… etc.) in the space. This paper presents SimArch, a multi-agent system for human moving path simulation. It involves a behavior model built by using a Markov Decision Process. The model simulates human mental states, target range detection, and collision prediction when agents are on the floor, in a particular small gallery, looking at an exhibit, or leaving the floor. It also models different kinds of human characteristics by assigning different transition probabilities. A modified weighted A* search algorithm quickly plans the sub-optimal path of the agents. In an experiment, SimArch takes a series of preprocessed floorplans as inputs, simulates the moving path, and outputs a density map for evaluation. The density map provides the prediction that how likely a person will occur in a location. A following discussion illustrates how architects can use the density map to improve their floorplan design.

## 1 Introduction

Can machines design or help human design? When an architect designs a building, human moving path is an important feature that he needs to consider. The path affects the basic elements (e.g. walls, columns, windows, furniture, stairs… etc.) placed in the building. Architects usually rely on their past experiences in visiting or designing the type of building to determine the possible human moving paths. But what if the architect doesn't have enough experiences? This paper focuses on the problem that how machines provide reasonable predictions and evaluations based on human behavior models.

There are several sub-problems. The first one is how to come up with a reasonable human behavior model. Human behaviors can be simplified to decision making processes. People make decisions every day in daily lives. They take actions according to current states. And their current states are determined by last states and actions. Ziebart et al. [8] used a soft-maximum Markov Decision Process with inverse reinforcement learning (maximum entropy inverse optimal control) to predict the decision uncertainty of pedestrians.

Second, how can the model represent human characteristics? There are a lot of different types of people. Some people like to wander around in the space while others like to look at their targets in detail. The model needs to represent the uncertainty of human characteristics. Therakomen [6] discussed the human steering behavior and used a geometry-based path planning algorithm to model wandering, passing-through, and obstacle avoiding behaviors.

Third, how does human plan paths to their targets and react when they are about to collide with others? Some people have maps in their hands or have visited the space before. So they know where exactly the targets are. While some people explores the space based on their intuition. The model needs to present a planning algorithm and collision prediction mechanism. Mukai et al. [3] developed a dynamic path setting for avoiding collision. Stipanović et al. [5] presented a collision avoidance function for robotic control. Yang et al. [7] used a neural network inspired by biology for dynamic collision-free path planning.

This paper presents a human moving path simulator for a museum exhibition space. It is a multi-agent system which takes a series of preprocessed floorplans as inputs, simulates moving paths as time goes on, and outputs an evaluation map for architecture design. The simulator uses a Markov Decision Process (MDP) to build the human behavior model. The states and actions in the MDP represent mental states (e.g. attracted by objects, boring…etc.) and actions (e.g. stay, move to a certain direction… etc.). The model assigns different transition probabilities to represent human characteristics. Some actions uses a modified weighted A* search algorithm for human path planning.

# 2 Method

## 2.1 The Preprocess of Floorplan Input

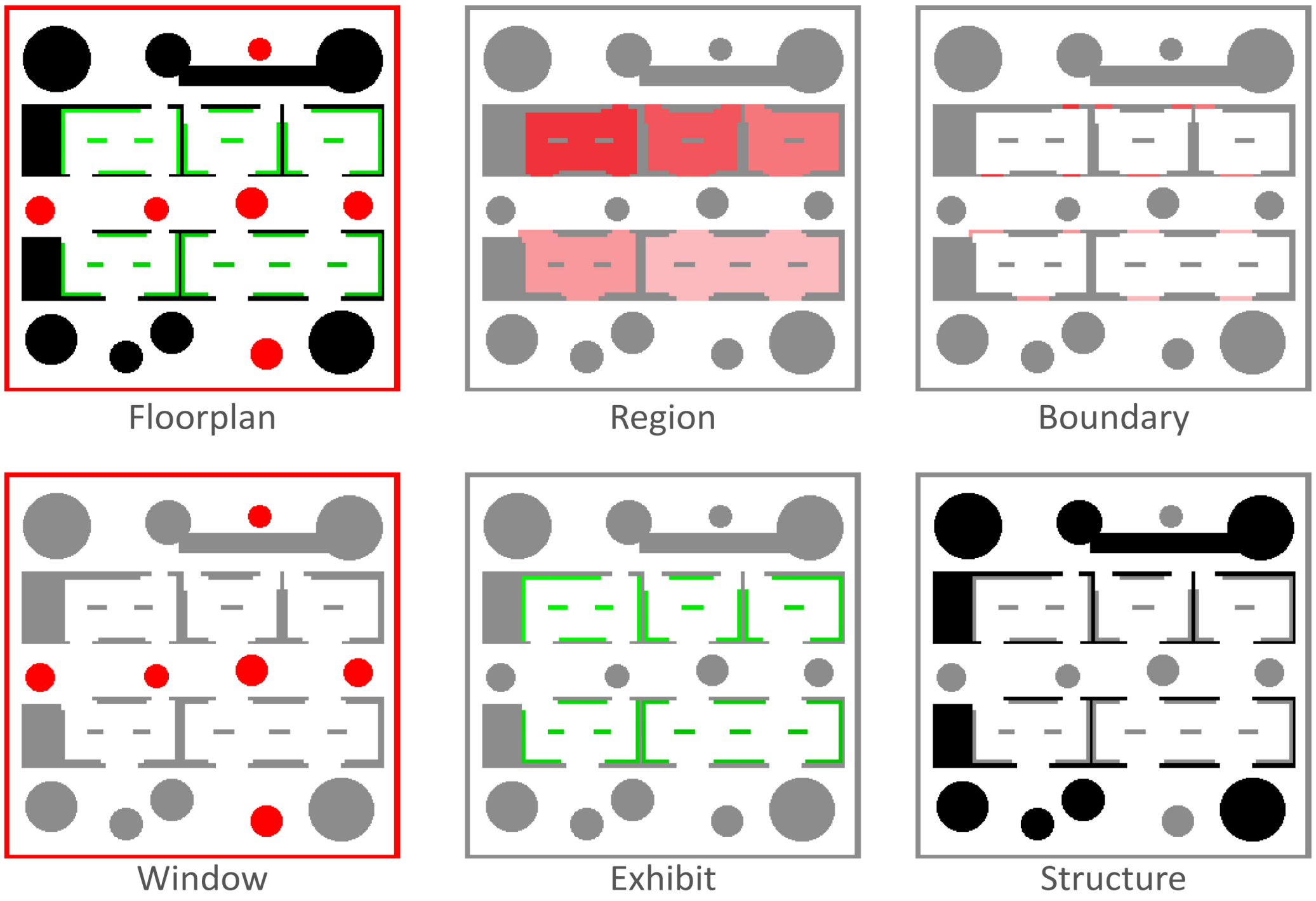

Figure 1: The input images of the simulator.

The floorplan is the 5th gallery of Sendai Mediatheque, Sendai city, Japan by Toyo Ito Architect. It is first down-sampled to 320 by 320 pixels. Each 1 by 1 pixel corresponds to about 7 by 7 square centimeters in the real space. Then 6 images are generated by assigning different colors to basic elements in the space, as shown in figure 1. There are 3 images for visible elements, 2 images for invisible elements, and 1 image for rendering. The 3 visible elements are “window”, “structure”, and “exhibit”. They represent glasses and light courts which are visually permeable, walls and columns which cannot be seen through, and art works. The 2 invisible elements are “region” and “boundary”, the area and the boundary (e.g. gates, doors, openings… etc.) of a particular small gallery. The “floorplan” image combines 3 visible elements. The simulator takes all 6 images as inputs. The starting and leaving point is at the elevator lobby on the bottom-left of the floorplan.

## 2.2 The Human Moving Path Model

Figure 3 shows the Markov Decision Process representing human moving behaviors. Generally speaking, in the left diagram of figure 2, the model contains three types of states: mental states (red), range detection (blue), and collision prediction (green). The mental states represent agents who are in the floor, in a particular gallery, interested by a target, or leaving the floor. Range detection states determine whether the agent is near or far away from the target. Collision prediction states prevent agents from colliding with each other. The middle diagram of figure 2 shows different functions for which the states are responsible. The orange, cyan, and purple color states deal with agents who are leaving, moving between the floor and the boundary, and moving between the boundary and the target. The right diagram of figure 2 explains how human characteristics are modeled. In states S2, S4, and S9, several probability parameters and a density function describe the uncertainty of human behaviors. By applying different parameters, the model simulates different human characteristics.

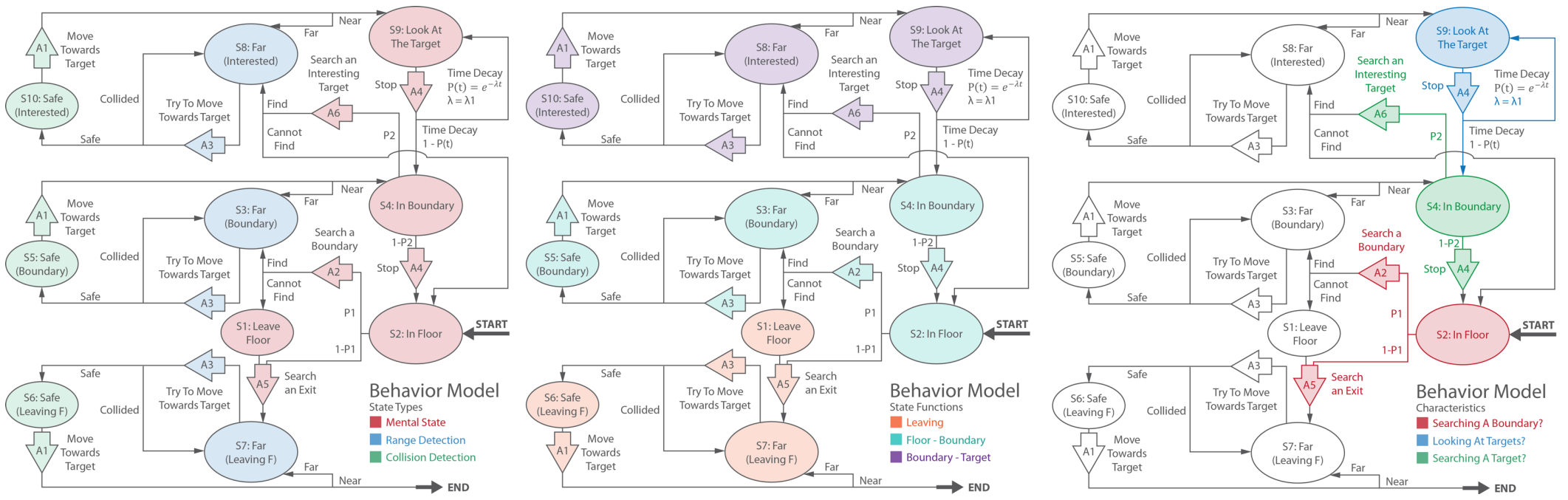

Figure 2: The general ideas of the model.

When an agent enters the exhibition space, the state starts from S2. S2 means that the agent is in the floor and can take two actions: searching a boundary of a small gallery (A2) with a P1 transition probability or searching an exit (A5) with a 1-P1 probability. The state then goes to S3 if the agent in S2 takes A2 and finds a boundary that hasn't been visited before, which means that the agent is far away from the boundary. The state goes to S1 if the agent cannot find a boundary, which means that the agent wants to leave the floor. On the other hand, the state goes to S7 if the agent in S2 takes A5 and finds an exit, which means that the agent is far away from the exit.

In S3, the agent takes A3, trying to move one step towards the target according to the path computed by A2. If the agent will collide with another one, the state goes back to S3. Otherwise the state goes to S5. S5 means that the agent can move safely. In S5, the agent takes A1, moving one step towards the target. Then the state goes to S4 or S3 if the agent is near or far from the boundary.

In S1 and S4, the model applies similar processes described in the last two paragraphs. But there is a large difference when the state starts from S4, goes through S8 and S10, and reaches S9. S9 means that the agent is near and looking at the exhibit in a particular small gallery. In S9, the agent stops (A4) at the same location. Then the model uses an exponential time decay function $P(t) = e^{-\lambda t}$ to compute the probability that the state goes to S4 or itself. If the $\lambda$ factor is large, P(t) decays faster, and vice versa. When the time goes on, the agent will have a lower probability to keep looking at the same exhibit and will have a higher probability to go to S4 and to search for another exhibit. If the state goes back to S4, it has a P2 probability to keep searching another exhibit and a 1-P2 probability to goes back to S2 and to search for another boundary.

A2, A5, and A6 use a modified weighted A* search algorithm to quickly find a sub-optimal moving path to a nearby target. The detail of weighted A* is beyond the range of this paper. Please refer to [4] for more information. The heuristic function is the Manhattan distance [1] between the current and the starting location. The modified version of weighted A* adds a uniformly distributed random noise to the heuristic function, as shown in the left image of figure 4. The darker the color, the smaller the heuristic value, and vice versa. The random noise prevents the agent from moving very closely to the obstacles. The middle image of figure 4 shows the expanded nodes. The red and yellow color indicates the CLOSE set and OPEN set. The white and black color indicates obstacles and empty area. The right image of figure 4 shows the sub-optimal path from the entrance to the boundary of a small gallery.

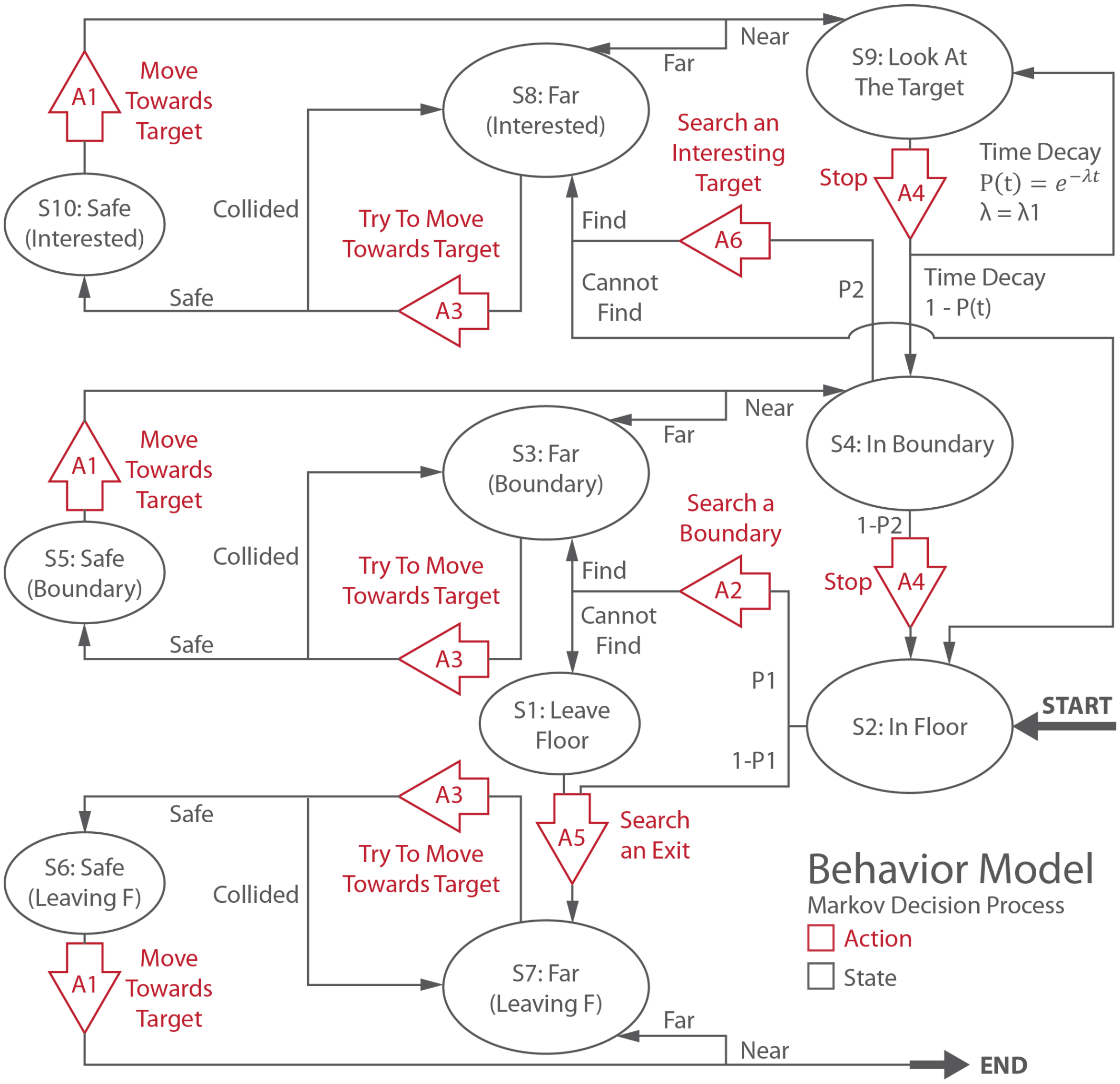

Figure 3: The behavior model of human moving paths.

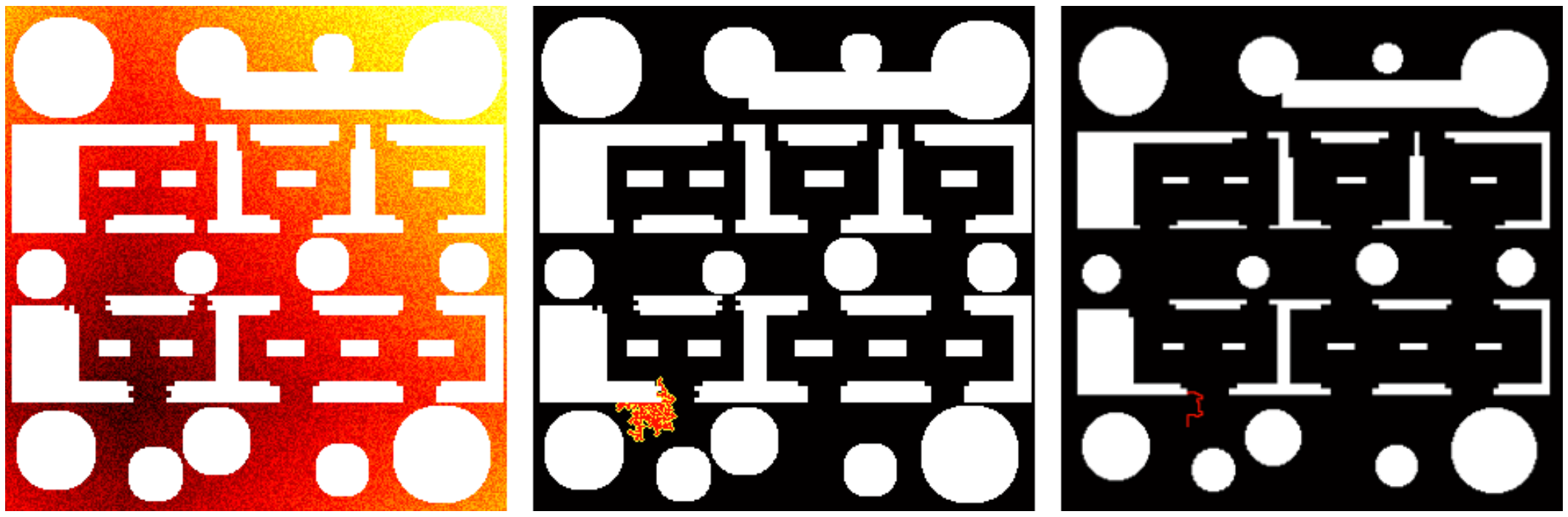

Figure 4: The modified weighted A* search algorithm.

## 3 Experiments

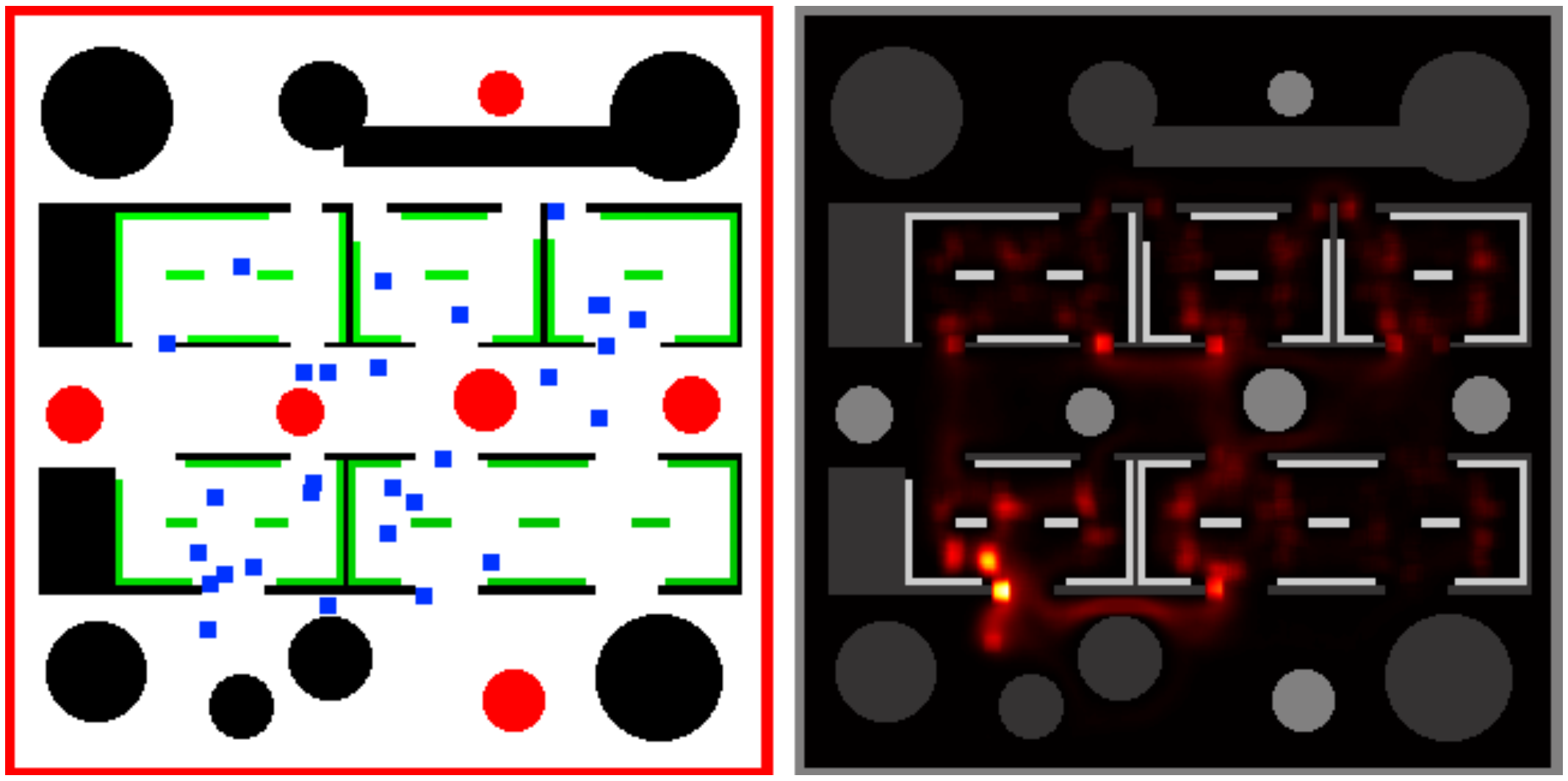

Figure 5: Left: the simulation. Right: the density map.

The left image of figure 5 is the simulation of the model. Agents start from the elevator lobby at the bottom-left of the floor. The black color objects in the map indicate the structure (e.g. walls and columns) of the building. The red color ones indicate windows and light courts. The green and blue color ones indicate the exhibits and the agents.

There are several parameters in the simulator. If the total number of agents in the floor is smaller than 30, the simulator adds 1 to 10 agents randomly and iteratively after a constant time span. The model randomly assigns the threshold between 10 and 20 pixels to define whether the agent is near or far away from its target. The model also randomly assigns values to P1, P2, and $\lambda 1$ to simulate different human characteristics. The lower and upper bounds of P1 and P2 are 0.5 and 1. The lower and upper bounds of $\lambda 1$ is 1/7000 and 1/500. In the modified weighted A* search algorithm, the weight used is 10. The variance of the random noise added in the heuristic function is 1000.

The right image of figure 5 is the density map which shows how likely that an agent will occur in the location. The brighter the pixel, the higher the probability. After an amount of time, the map converges. Density map is an important evaluation of the floorplan. For instance, in figure 6, the left-most image shows that the density on the boundaries facing the entrance is higher. The architect might need to increase the size of these openings. The left-middle image illustrates that the density in the gallery near the entrance is higher. The size of the gallery might need to be increased. In the right-middle image, there are a lot of paths inside the red box. The architect can design a resting area here for people walking by. The right-most image indicates the way to increase the user experience. Placing eye-catching exhibits at some locations can strengthen the connectivity between small galleries.

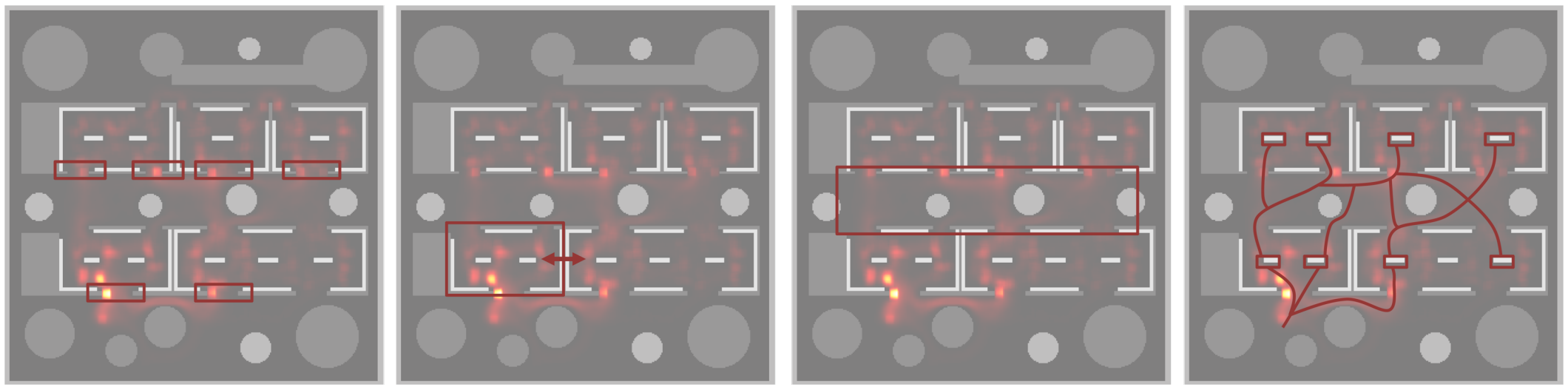

Figure 6: The evaluation of the floorplan using density map.

## 4 Discussion and Future Work

There are several problems and possible improvements of this model. First, the model doesn't involve the human characteristics in figure 7. Some people prefer walking to and stay in quiet spaces while others prefer crowd spaces. Some people love following other people while others doesn't. Some people like exploring the space and walk toward the target in their vision fields while others like to look at maps.

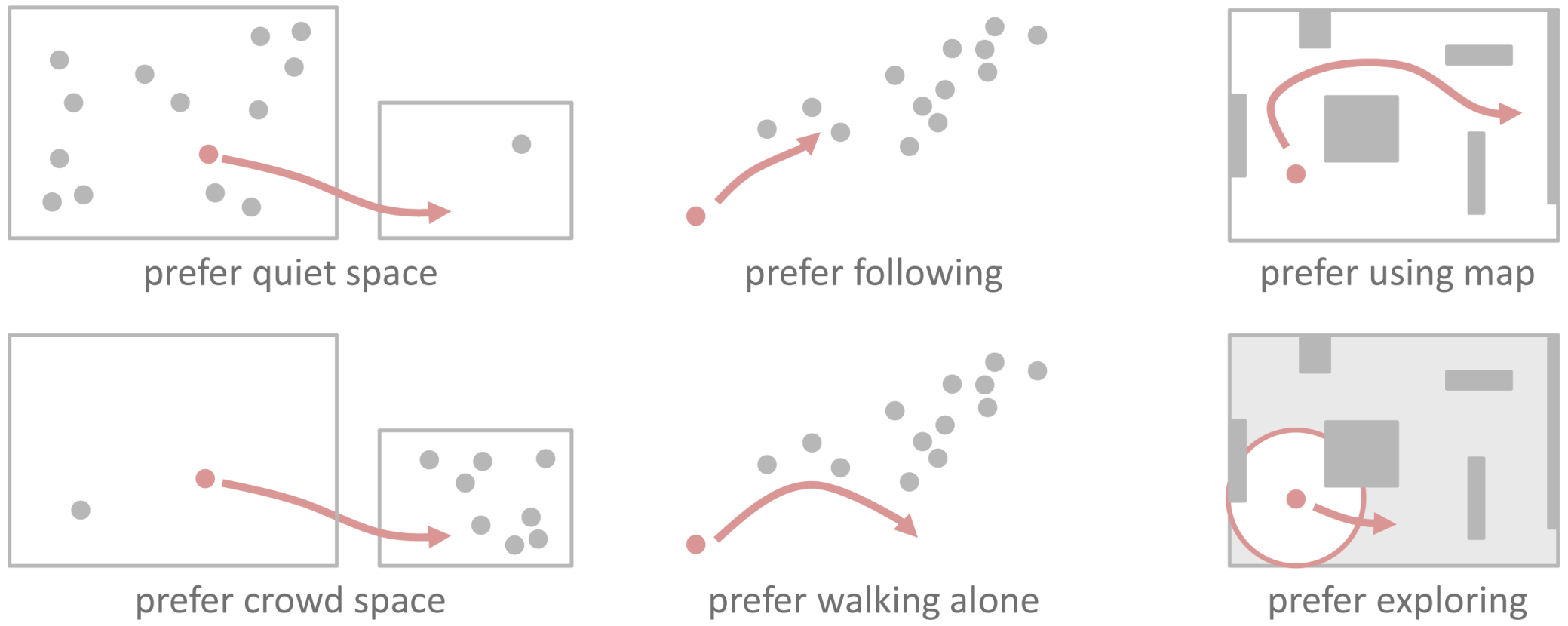

Figure 7: More human characteristics.

Second, there is a bug in the model for collision prediction. When an agent tries to move toward the target and figures out that it will collide with another one, the agent will wait at the same location until safe. The bug is that two agents will halt when they are waiting for each other. The model needs a better collision prediction mechanism and an algorithm that can repair the planning path dynamically (e.g. anytime repairing A* search [2]). But in fact, the bug seldom shows up in the simulation because of the random noise added in the heuristic function.

Third, in architecture design, architects are not only interested in simulating one floorplan but also the entire building. The model now only considers the horizontal relationship (floorplan) in the building. In real situation, vertical relationships (elevation) also affect human moving paths. For instance, a space (e.g. a high-ceiling atrium) that has a broader view of multiple floors is more likely to attract people.

## 5 Conclusion

This paper presents a human moving path simulator, SimArch, for architecture design. SimArch uses a Markov Decision Process as the behavior model. The model involves human mental states, target range detection, and collision prediction. It also models different human characteristics by assigning different transition probabilities. A modified weighted A* search algorithm quickly computes the sub-optimal moving path. SimArch takes a series of preprocessed gallery floorplans as inputs, uses the behavior model for simulation, and outputs a density map for evaluation which helps architects to improve the floorplan design.